\documentclass{article}

\PassOptionsToPackage{numbers, compress}{natbib}

\usepackage[preprint]{neurips_2025}

\usepackage[utf8]{inputenc}   
\usepackage[T1]{fontenc}      
\usepackage{CJKutf8}

\usepackage{url}

\usepackage{amsmath, amsfonts, amssymb}  
\usepackage{bbding}

\usepackage{graphicx}        
\usepackage{wrapfig}          
\usepackage{float}          
\usepackage{subfigure}        
\usepackage{capt-of}         

\usepackage{booktabs}        
\usepackage{longtable}        
\usepackage{ltablex}          
\usepackage{pdflscape}       
\usepackage{array}            
\usepackage{multirow}         
\usepackage{arydshln}

\usepackage{microtype}       
\usepackage{nicefrac}         
\usepackage{xspace}

\usepackage{xcolor}          
\usepackage{colortbl}         
\usepackage[most]{tcolorbox} 

\usepackage{makecell}
\usepackage{pifont}           

\keepXColumns

\newcommand{\benchmark}{\textsc{UI-Nexus}\xspace}
\newcommand{\offlinebench}{\textsc{UI-Nexus-Anchor}\xspace}
\newcommand{\agent}{\textsc{Agent-Nexus}\xspace}
\newcommand{\xmark}{\textcolor{red}{\ensuremath{\times}}}
\newcommand{\cmark}{\textcolor[rgb]{0,0.51,0}{\checkmark}}

\title{Atomic-to-Compositional Generalization for Mobile Agents with A New Benchmark and Scheduling System}

\author{
    Yuan Guo\textsuperscript{\rm 1,\rm 2}\thanks{Work done during Yuan's internship at Langboat Technology} \quad
    Tingjia Miao\textsuperscript{\rm 1} \\
    \textbf{Zheng Wu\textsuperscript{\rm 1} \quad
    Pengzhou Cheng\textsuperscript{\rm 1} \quad
    Ming Zhou\textsuperscript{\rm 2} \quad
    Zhuosheng Zhang\textsuperscript{\rm 1}\thanks{Corresponding author}}\\
    \textsuperscript{\rm 1}Shanghai Jiao Tong University \quad \textsuperscript{\rm 2}Langboat Technology \\
    \texttt{yuanguo2004@gmail.com, zhangzs@sjtu.edu.cn}
}

\begin{document}
\maketitle

\begin{figure}[h!]
\centering
\vspace{-10mm}
\subfigure[The atomic-to-compositional generalization gap. In the task subset, the agents demonstrates significant performance gap between directly given the compositional task instructions and given manually decomposed atomic task instructions to conduct subtasks separately. Our \agent scheduling system significantly covers the generalization gap.]{\includegraphics[width=0.46\linewidth]{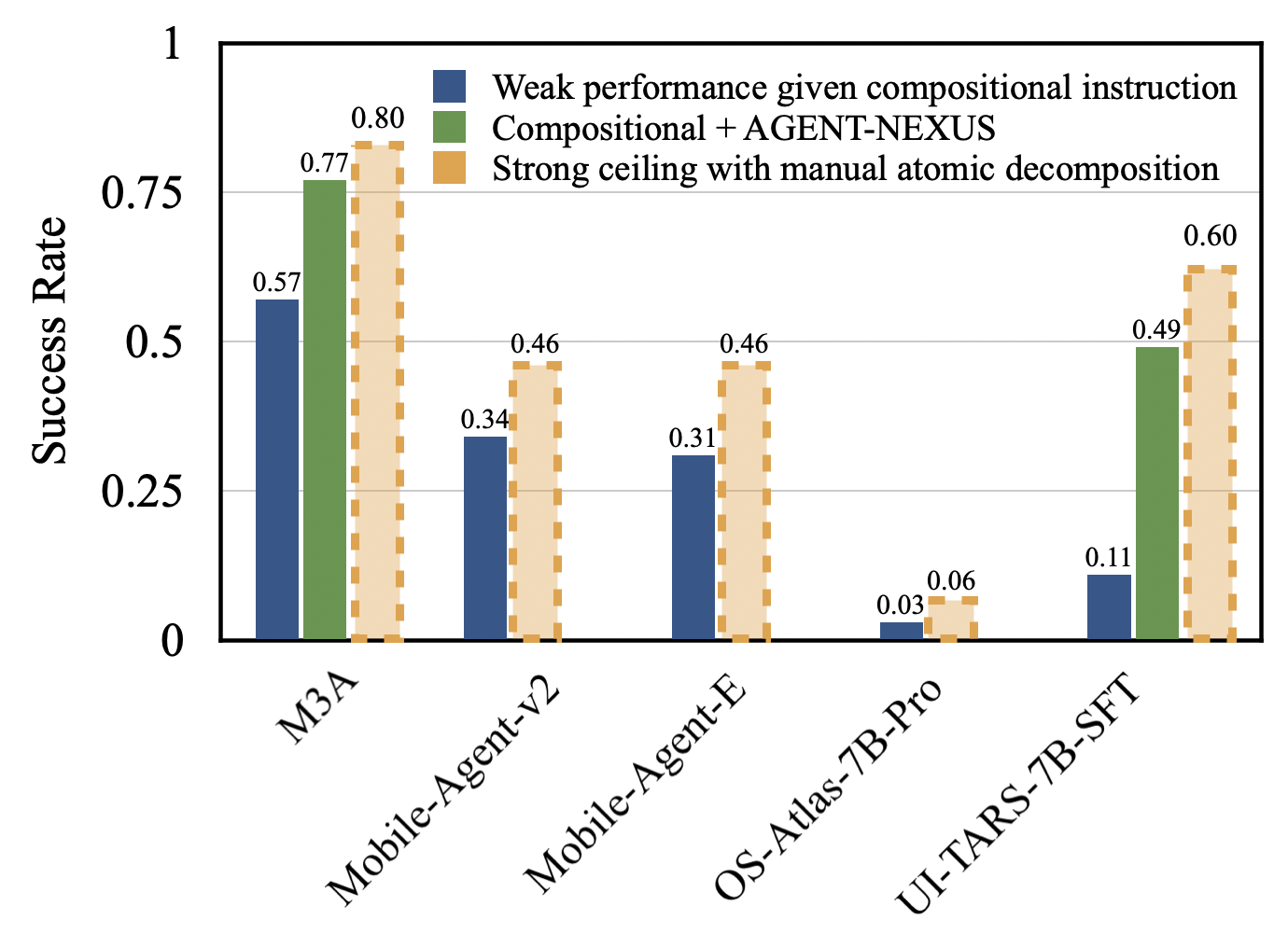}}
\hfill
\subfigure[Task performance and token cost comparison. \agent brings about significant performance improvement with controllable sacrifice of inference cost on the compositional tasks in \offlinebench benchmark.]{\includegraphics[width=0.48\linewidth]{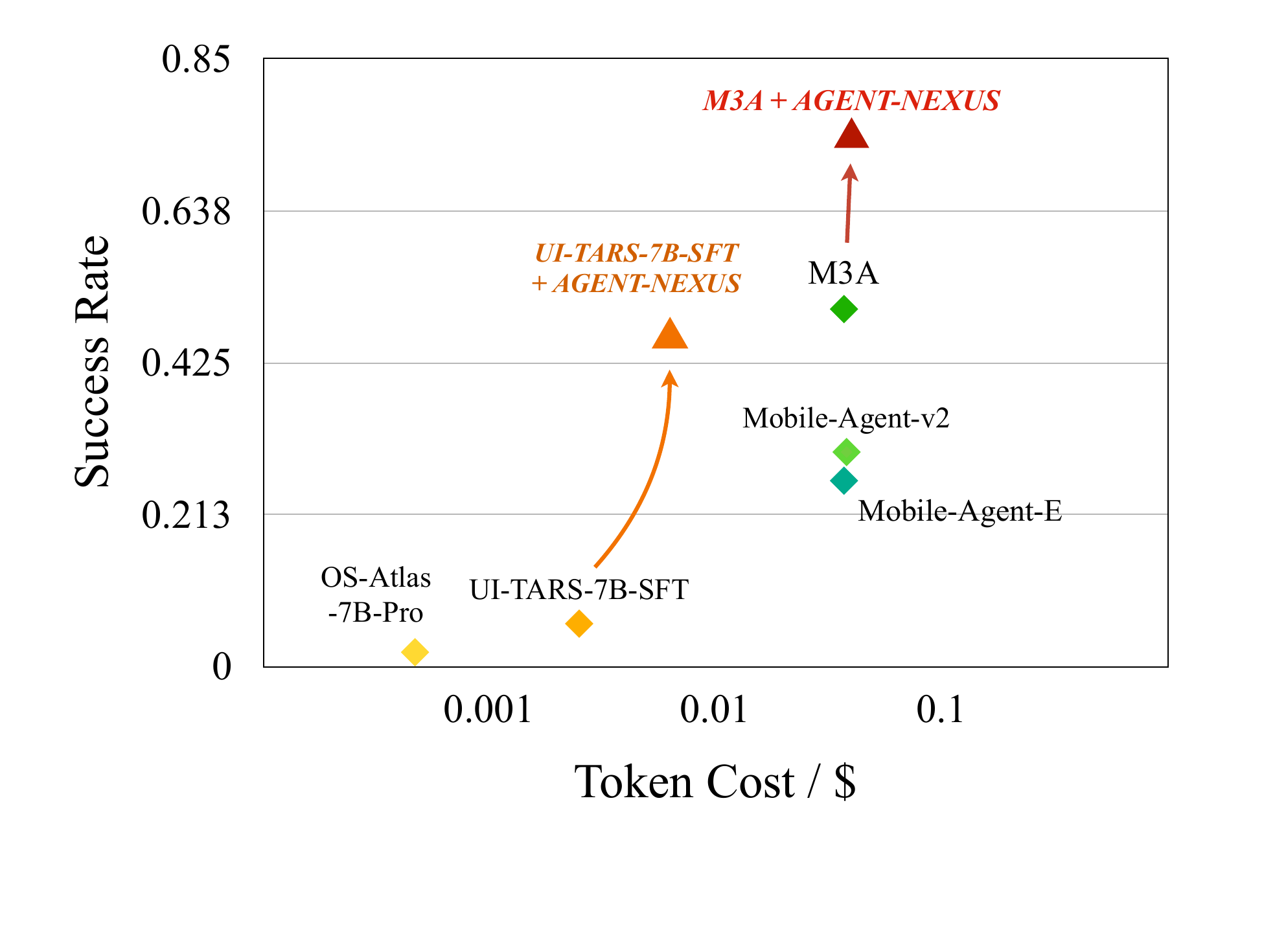}}
\caption{Visualization of the atomic-to-compositional generalization issue of mobile agents.}
\label{fig:comparison}
\end{figure}

\begin{abstract}

Autonomous agents powered by multimodal large language models have been developed to facilitate task execution on mobile devices.
However, prior work has predominantly focused on atomic tasks---such as shot-chain execution tasks and single-screen grounding tasks---while overlooking the generalization to compositional tasks, which are indispensable for real-world applications.
This work introduces \benchmark, a comprehensive benchmark designed to evaluate mobile agents on three categories of compositional operations: \textit{Simple Concatenation}, \textit{Context Transition}, and \textit{Deep Dive}. 
\benchmark supports interactive evaluation in 20 fully controllable local utility app environments, as well as 30 online Chinese and English service apps.
It comprises 100 interactive task templates with an average optimal step count of 14.05.
Experimental results across a range of mobile agents with agentic workflow or agent-as-a-model show that \benchmark presents significant challenges. 
Specifically, existing agents generally struggle to balance performance and efficiency, exhibiting representative failure modes such as under-execution, over-execution, and attention drift, causing visible atomic-to-compositional generalization gap.
Inspired by these findings, we propose \agent, a lightweight and efficient scheduling system to tackle compositional mobile tasks. 
\agent extrapolates the abilities of existing mobile agents by dynamically decomposing long-horizon tasks to a series of self-contained atomic subtasks. \agent achieves 24\% to 40\% task success rate  improvement for existing mobile agents on compositional operation tasks within the \benchmark benchmark without significantly sacrificing inference overhead. 
The demo video, dataset, and code are available on the project page \url{https://ui-nexus.github.io}.

\end{abstract}

\section{Introduction}
\label{sec:introduction}

Recent advances in multimodal large language models (MLLM) have spurred the development of AI agents capable of autonomous operation on intelligent devices~\citep{zhang2024large,liu2025llm,nguyen2024gui,hu2024agents,wang2024gui}. 
These agents can interact with digital systems and navigate through diverse applications to accomplish tasks.
This progress marks a significant leap forward in human-computer interaction and automation.
The latest studies have shown that these agents exhibit proactive planning, reasoning, reflection, and adaptation capabilities~\citep{bai2024digirl,wang2024distrl,liu2024autoglm,qin2025ui,jiang2025appagentx}, enabling them to handle a wide range of tasks effectively.
Popular applications include action grounding~\citep{wu2024atlas, gou2024navigating}---i.e., mapping natural language instructions to specific screen coordinates---as well as information seeking~\citep{chai2025a3, rawles2024androidworld, xu2024androidlab} and end-to-end task completion~\citep{rawles2024androidinthewild, zhang2023you, you2024ferret}. One important branch of these device-using agents is mobile agents (i.e. the agents that interact with mobile devices like smartphones), which play a pivotal role in bringing AI autonomy to the most pervasive computing platform in daily life.

\begin{figure*}[t]
\scriptsize
    \centering
    \includegraphics[width=1.0\textwidth]{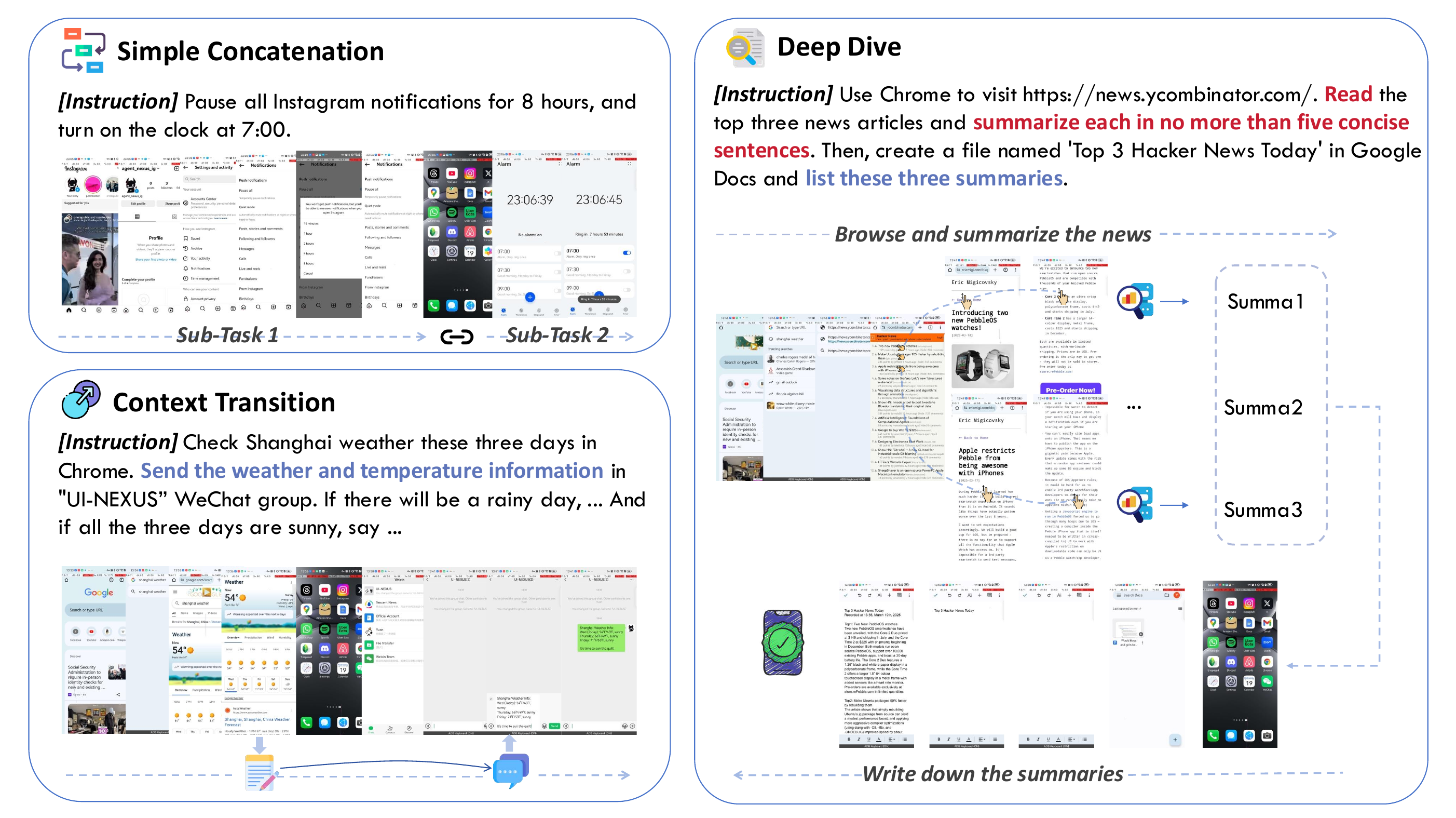}
    \caption{Illustration of the composition type taxonomy in \benchmark benchmark. According to the subtask dependency structure, we identify three types of compositional tasks: (i) Simple Concatenation refers to cases where the compositional instructions are direct concatenations of atomic subtasks; (ii) Context Transition refers to cases where the atomic subtasks require explicit transition of shared context; (iii) Deep Dive refers to cases where in-depth analysis, like information aggregation and logical reasoning is required for transitions of atomic operation tasks.}
    \label{fig:taxonomy}
    \vspace{-5mm}
\end{figure*}

Despite rapid progress, most existing research and evaluations focus on atomic tasks whose intents are often generic and specific, such as short-chain execution tasks~\citep{rawles2024androidinthewild,li2024effects} (e.g., \textit{`Look up the best rated coffee maker'}) and single-screen grounding tasks (e.g., \textit{`Click the submit button'})~\citep{cheng2024seeclick, gou2024navigating}. However, real-world scenarios often involve compositional tasks with compositional intents or complex dependencies.
As illustrated in Figure \ref{fig:taxonomy}, practical user instructions frequently require concatenated operation sequences, conditional branching based on information retrieval results, and in-depth information processing such as aggregation and analysis.

These compositional tasks bring about distinctive requirements such as long-horizon progress management, intermediate information transition, and seamless integration of device operations and general reasoning. Even with trained device interaction logic, agents may still face challenges and exhibit failure modes such as over-execution (e.g., failing to propagate necessary intermediate information, causing subsequent subtasks to be instantiated with fabricated or irrelevant content), under-execution (e.g., switching between multiple applications without solidly completing subtasks one by one), context confusion (e.g., fulfilling requirements for app B while actually interacting with app A), and attention drift (e.g., neglecting some task requirements or subtasks altogether).\footnote{More detailed categorization and examples of these failure modes are elaborated in Appendix~\ref{appendix:error-analysis}.} However, existing agents and evaluation benchmarks rarely investigate these sophisticated dependencies and compositional requirements.

\definecolor{c5}{RGB}{51,114,202}
\definecolor{c6}{RGB}{121,43,166}

To fill this crucial gap, we systematically investigate atomic-to-compositional generalization in mobile agents. 
We specifically focus on instruction-level atomic-to-compositional generalization and define \textbf{atomic subtasks} as self-contained, straightforward operation tasks.
To investigate into how well the existing mobile agents can generalize to the sophisticated composition of these subtasks, we introduce \benchmark, a benchmark that systematically evaluates three types of task compositions, namely \textit{Simple Concatenation}, \textit{Context Transition}, and \textit{Deep Dive}---based on subtask dependency structures. \benchmark features bilingual (Chinese and English) instructions and includes 20 local utility apps along with 30 online service apps. 
To support scalable evaluation and development, we present a modular and extensible infrastructure with abstracted interfaces for device state control, agent framework integration, and task evaluation.
We conduct extensive evaluations on 5 mainstream mobile agents following agentic workflow or agent-as-a-model implementations. 
Experimental results show that \benchmark poses substantial challenges to current mobile agents. 
These agents exhibit representative failure modes facing task composition caused by insufficient progress management and context overflow, with some agents also incurring substantial computational redundancy. 
In light of these findings, we propose \agent, a lightweight and scalable scheduling system for tackling compositional mobile tasks. 
\agent effectively extrapolates the abilities of existing mobile agents by dynamically decomposing the long-horizon tasks to a series of self-contained atomic subtasks.  
\agent achieves remarkable improvement on compositional operation tasks within the \benchmark benchmark without significantly sacrificing inference overhead. 
\agent shows to be helpful in handling unstructured expressions and sophisticated dependencies within user queries. 

\section{Related Work}

\subsection{Mobile Agent}
\label{sec:related-gui-agent}

Mobile agents refers to the autonomous AI agents that can interact with mobile devices to complete tasks. They can be developed by API-based~\citep{deng2024mobile} operations or UI-based operations~\citep{zhang2023appagent, wen2024autodroid}. Due to the diversity and lack of api interface of mobile device applications, most of the recent mobile agents rely on human-like Graphical User Interface (GUI) operations for task fulfillment. All the mobile agents studied in this paper are implemented as MLLM-powered GUI agents.
Following SPA-Bench~\citep{chen2024spa}, they can be further categorized into agentic workflow (relies on off-the-shelf models and modular designs to support agentic functionality) and agent-as-a-model (where fine-tuned or pre-trained (M)LLMs are customised for agentic tasks).

$\bullet$ \textbf{Agentic workflow} typically exploit the strong general abilities of powerful backbones like GPT-4o to build versatile systems. 
Various design techniques have been explored, including environment exploration~\citep{zhang2023appagent, li2024appagent}, multi-agent collaboration~\citep{wang2024mobile1, wang2024mobile2, wang2025mobilee, zhang2024ufo}, planning-grounding decouple~\citep{zheng2024gpt}, self-reflection~\citep{rawles2024androidworld, liu2025infiguiagent}, experience utilization~\citep{agashe2024agent, jiang2025appagentx, wang2024agent}.
Complementary strategies like screen parser tools~\citep{lu2024omniparser, rawles2024androidinthewild} and Set-of-Marks prompting~\citep{yang2023set, yan2023gpt} are also applied to improve the multimodal environment perception.

$\bullet$ \textbf{Agent-as-a-model} train (M)LLM backbones on GUI domain-specific data for GUI skill acquisition. Through task-specific model architecture design~\citep{lu2024gui, lin2024showui, huang2025spiritsight}, pre-training tasks like action grounding and element reference~\citep{you2024ferret, gou2024navigating}, action fine-tuning on static trajectories~\citep{zhang2023you,zhang2024android} and preference learning strategies~\citep{bai2024digirl,wang2024distrl,qi2024webrl,putta2024agent}, open-source models have been adapted to the specific scenario of mobile device GUI navigation. 
Recently, high-quality training data at scale and advanced training strategies have catalyzed the emergence of various strong GUI foundation action models like OS-Atlas~\citep{wu2024atlas}, Aguvis~\citep{xu2024aguvis}, and UI-TARS~\citep{qin2025ui} that achieve competitive performance on many well-established interactive benchmarks like AndroidWorld~\citep{rawles2024androidworld}.

\subsection{GUI Agent Evaluation}
\label{2_2}

Currently, there are three main types of benchmarks to evaluate abilities of GUI agents: domain skill evaluation, offline action prediction evaluation and end-to-end task fulfillment evaluation.

$\bullet$ \textbf{Domain skill benchmarks} evaluate models and agents on GUI-specific tasks. These benchmarks include fundamental domain-specific skills like visual grounding~\citep{cheng2024seeclick, wu2024atlas, li2024screenspot-pro, fan2025gui}, interface comprehension~\citep{hsiao2022screenqa, chen2021websrc, liu2024visualwebbench} and information processing~\citep{lu2024weblinx}.

$\bullet$ \textbf{Offline action prediction benchmarks} test GUI agents' step-level action accuracy against static golden trajectories. Agents predict the next action given task details, current observation and history, evaluated against annotated golden actions. Benchmarks like Android-in-the-Wild~\citep{rawles2024androidinthewild}, AndroidControl~\citep{li2024effects}, Meta-GUI~\citep{sun2022meta} and Mind2Web~\citep{deng2023mind2web} provide large-scale annotated GUI data split into training and test sets for offline evaluation.

$\bullet$ \textbf{End-to-end task fulfillment benchmarks} put agents in an interactive environment like Android Emulator, virtual machines and web environment, and let the agent fulfill the user goal end-to-end. Representative end-to-end GUI operation benchmarks include MobileAgentBench~\citep{wang2024mobileagentbench}, SPA-bench~\citep{chen2024spa}, AndroidLab~\citep{xu2024androidlab}, A3~\citep{chai2025a3} for mobile device tasks and OSWorld~\citep{xie2025osworld}, WebArena~\citep{zhou2023webarena}, WebCanvas~\citep{pan2024webcanvas}, Windows Agent Arena~\citep{bonatti2024windows} and WorkArena~\citep{drouin2024workarena} for computer use.

Several related benchmarks have contributed to progress in addressing compositional GUI tasks by introducing tasks with cross-app sequences and varying compositional complexity~\citep{lu2024gui,chen2024spa,zhang2024ufo,rawles2024androidworld,wang2024mobile2,wang2025mobilee,liu2025pc}.

Yet compositional tasks in these benchmarks are typically sporadic and simplified into concatenated cross-app sequences, lacking systematic definitions and thorough analyses. Moreover, tasks integrating detailed execution sequences with in-depth information processing, such as expense-statistic tasks that require non-trivial general reasoning during device interactions, remain underexplored.
Considering the dynamic nature of devices, establishing an interactive, controllable benchmark tailored for compositional operations is essential for subsequent scalable research on long-horizon cognitive GUI agents.

\section{The \benchmark Benchmark}
\subsection{Overview of \benchmark}
\label{sec:benchmark_overview}

\begin{figure*}[t]
\scriptsize
    \centering
    \includegraphics[width=0.95\textwidth]{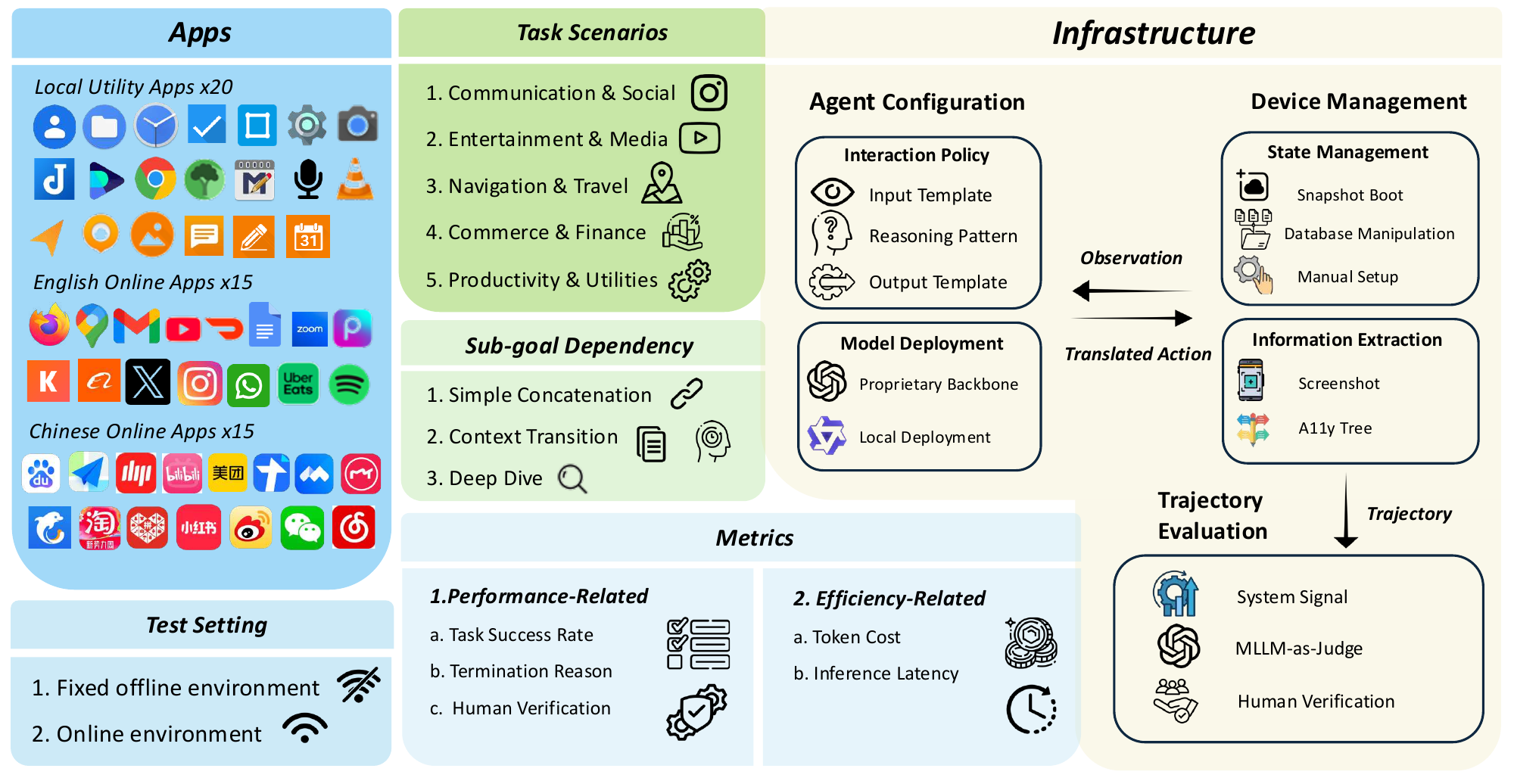}
    \caption{Overview of \benchmark. It features comprehensive coverage of applications and task scenarios, systematic analysis of subtask dependencies, diverse evaluation metrics, and supports both fully reproducible offline evaluations and real-world online tests. UI-NEXUS is built upon our unified plug-and-play framework that seamlessly integrates heterogeneous agents and devices.}
    \label{fig:benchmark_overview}
    \vspace{-5mm}
\end{figure*}

We introduce \benchmark, a comprehensive benchmark to comprehensively evaluate the performance of mobile agents on compositional operation tasks. 
The characteristics of the \benchmark benchmark is presented in Figure \ref{fig:benchmark_overview}.

\benchmark is built upon our unified plug-and-play infrastructure for managing heterogeneous mobile GUI agents. 
Our infrastructure is structured into three interconnected high-level modules: Agent Configuration, Device Management, and Trajectory Evaluation, providing a standardized foundation for efficient deployment and evaluation of diverse mobile GUI agents. 

\definecolor{c5}{RGB}{51,114,202}
\definecolor{c6}{RGB}{121,43,166}

\benchmark also features systematic and multi-discipline investigation into compositional GUI operation tasks. The task instructions are organized in three kinds of subtask dependency structures (Simple Concatenation, Context Transition and Deep Dive), and boasts a wide coverage of 5 kinds of use scenarios in 20 local utility apps and 30 English and Chinese online service apps. With meticulously designed performance-related and efficiency-related evaluation metrics, \benchmark enables a fine-grained evaluation of the performance of different mobile agents on compositional tasks. 
Inspired by AndroidWorld \citep{rawles2024androidworld}, we include a fully controllable, reproducible, and parameterized subset, \offlinebench in \benchmark, to provide a standardized platform for developing and comparing compositional mobile-automation methods. \offlinebench focuses on the 20 local utility tasks whose internal states can be precisely manipulated through database and file-system operations.

\subsection{Task Formulation}

In this section, we formalize the definition of device-use task composition and evaluation of mobile agents through a structured task formulation.

$\bullet$ \textbf{Device Use Task} Following previous work~\citep{wu2025vsc,bai2024digirl} and classical definitions~\citep{puterman1990markov}, we model the task of device use as a finite-horizon, goal-conditioned Markov Decision Process (MDP). Formally, the MDP is represented as a tuple $\langle \mathcal{G}, \mathcal{S}, \mathcal{A}, \mathcal{T}, \mathcal{R}, H \rangle$, where $\mathcal{G}$ denotes the set of possible goals, $\mathcal{S}$ represents the state space 
which can consist of  visual screenshots, a11y trees etc
, and $\mathcal{A}$ defines the set of executable actions (e.g., typing text, pressing buttons, or swiping screens). The state transition dynamics are governed by the function $\mathcal{T}: \mathcal{S} \times \mathcal{A} \times \mathcal{S} \rightarrow [0, 1]$, specifying the probability of transitioning to a new state given a current state-action pair. $\mathcal{R}$ is the reward function assigning a reward $r_t = 1$ upon successful achievement of a predefined goal $g \in \mathcal{G}$ at timestep $t$, and $r_t = 0$ otherwise. The horizon $H$ sets a finite limit on the maximum number of timesteps for each episode.

At each timestep $t$, given a policy $\pi$ provided by the agent under evaluation, the agent observes the current state $s_t \in \mathcal{S}$ and selects an action $a_t \in \mathcal{A}$ according to $\pi: \mathcal{S} \times \mathcal{G} \times \mathcal{A} \rightarrow [0, 1]$, conditioned on the current goal $g$. The benchmark's objective is to assess the effectiveness of the provided policy $\pi$ in consistently achieving all goals from $\mathcal{G}$ within the stipulated horizon $H$.

$\bullet$ \textbf{Atomic Subtask} In this work, we define an atomic subtask $a$ as a self-contained instruction unit that encapsulates complete semantic meaning in the mobile environment, without further scenario-specific decomposition to specified action sequences. Formally, we denote it as \(a \triangleq \langle c, p, e \rangle\), where $c$ denotes the command or intention, $p$ denotes necessary parameters, and $e$ denotes the execution environment.
In this paper, we focus on instruction-level atomic decomposition. For instance, \textit{``Search tomorrow's Shanghai weather using Chrome''} is atomic, and we don't consider further scenario knowledge-based decomposition like  \textit{``Click search button''}.

$\bullet$ \textbf{Compositional Task} 
A compositional mobile task is defined as a set of atomic subtasks with dependency relations, represented as \(\mathcal{T}_{comp} \triangleq \langle \mathcal{A}_{sub}, \mathcal{D} \rangle\), where \(\mathcal{A}_{sub}\) denotes a finite set of atomic subtasks, and \(\mathcal{D}\) encodes the dependency structure among them. Based on the interaction patterns among subtasks, we categorize compositional tasks into three types:

- \textit{Simple Concatenation}: A sequence of subtasks \((a_1, a_2, \ldots, a_n)\) that are executed independently without cross-subtask state dependencies. The agent applies the policy \(\pi\) to each subtask individually, conditioned on its local goal \(g_i\).

- \textit{Context Transition}: A composition where certain subtasks depend on the outcome states of others. Formally, if there exists \((a_i, a_j) \in \mathcal{D}\), then the input to subtask \(a_j\) is the state \(s_i'\) resulting from executing \(a_i\) under \(\pi\) and transition dynamics \(\mathcal{T}\).

- \textit{Deep Dive}: A special case of context transition where intermediate reasoning beyond direct state transitions is required. We introduce a reasoning function \(\mathcal{F}: \mathcal{S} \times \mathcal{G} \rightarrow \mathcal{I}\), mapping the current state and goal to latent intermediate information \(\mathcal{I}\), which further guides action selection under \(\pi\).

\begin{table}[t]
    \centering
    \small
    \setlength{\tabcolsep}{4pt}  
    \begin{tabular}{l l c c c c c c}
        \toprule
        \textbf{Dataset} & \textbf{Platform} &
        \makecell{\# Apps\\or Sites} &
        \makecell{\# Avg\\Steps} &
        \makecell{\# Task\\Templates} &
        \makecell{Experiment \\ Environment} &
        \makecell{Task \\ Scalability} &
        \makecell{Compositional \\ Type} \\
        \midrule
        Mind2Web~\citep{deng2023mind2web}        & Web      & 137 & 7.3      & 2350   & \xmark & \xmark & \xmark \\
        WebArena~\citep{zhou2023webarena}        & Web      & 6   & --       & 241    & \cmark & \cmark & \xmark \\
        OSWorld~\citep{xie2025osworld}           & Desktop  & 9   & $\sim$15 & 369    & \xmark & \cmark & \xmark \\
        GUI-Odyssey~\citep{lu2024gui}            & Mobile   & 201 & 15.4     & --     & \xmark & \cmark & \xmark \\
        AndroidControl~\citep{li2024effects}     & Android  & 833 & 5.5      & 14548  & \xmark & \xmark & \xmark \\
        AndroidLab~\citep{xu2024androidlab}      & Android  & 9   & --       & 138    & \cmark & \xmark & \xmark \\
        Mobile-Eval-E~\citep{wang2025mobilee}    & Android  & 15  & 14.56    & 25     & \xmark & \xmark & \xmark \\
        AndroidWorld~\citep{rawles2024androidworld} & Android & 20  & 8.41     & 116    & \cmark & \cmark & \xmark \\
        \benchmark                                 & Android  & 50  & 14.05    & 100    & \cmark & \cmark & \cmark \\
        \bottomrule
    \end{tabular}
    \vspace{0.3cm}
    \caption{Comparison of representative GUI benchmarks. 
    The \textit{Env} column indicates whether an interactive testing environment is available (offline benchmarks such as \emph{AndroidControl} provide only static trajectories, whereas end-to-end interactive suites like \emph{Mobile-Eval-E} supply task instructions without an environment). 
    \textit{Scalable} denotes whether the task set can be readily expanded to generate additional task variants and instructions by easy techniques like changing subjects in the task templates.}
    \label{tab:comparison}
\end{table}

\subsection{Benchmark Construction}
\label{sec:benchmark_construction}

Our benchmark is built upon a modular infrastructure and a carefully curated task set. The infrastructure consists of three core components: \textbf{Device Management}, \textbf{Agent Configuration}, and \textbf{Trajectory Evaluation}. These modules support streamlined device control (e.g., emulator snapshots, offline database editing), unified execution interfaces, and trajectory evaluation with hybrid metrics.

For task collection, we identify 50 widely-used mobile apps, covering both local utilities and online services, across five common usage scenarios. We adopt the three subtask dependency structures, and construct 20 seed tasks with diverse logic structures (e.g., sequential, parallel, conditional, and hierarchical). GPT-4o and GPT-o1 are employed to expand the task set, followed by human refinement to ensure task quality and type balance.
Full implementation details, prompt templates, and data generation procedures are provided in Appendix~\ref{appendix:data-collection}.

\subsection{Evaluation Metrics}

We design fine-grained metrics for \benchmark benchmark. The metrics can be generally divided into performance-related metrics and efficiency-related metrics.
More details about evaluation metrics are listed in Appendix \ref{appendix:evaluation_metrics}.

\textbf{Performance-Related Metrics}. 
(1) \textbf{Task Success Rate} — The ratio of successfully completed tasks. In offline evaluation, we follow AndroidWorld~\citep{rawles2024androidworld} to implement automatic reward signals; for online evaluation, we follow previous work to apply MLLM-as-a-Judge~\citep{chai2025a3, chen2024spa} and human verification~\citep{wang2025mobilee} for accurate verification across diverse environments. 
(2) \textbf{Termination Reason} — The termination reasons are categorized as follows: 
(i) Successful Termination: The task is terminated after fulfilling all requirements; 
(ii) Premature Termination: The task is terminated without exactly meeting the requirements; 
(iii) Step Budget Exceeded: The task exceeds the predefined maximum number of steps; 
(iv) Deemed Impossible: The agent determines the task cannot be completed; 
(v) Collapse: Malformed output templates cause the failure of execution.

\textbf{Efficiency-Related Metrics}. 
(1) \textbf{Inference Cost} — The average monetary cost of model inference per step. 
(2) \textbf{Inference Latency} — The average inference time per step. 

\subsection{Benchmark Statistics}

\begin{figure*}[t]
    \centering
    \includegraphics[width=0.95\textwidth]{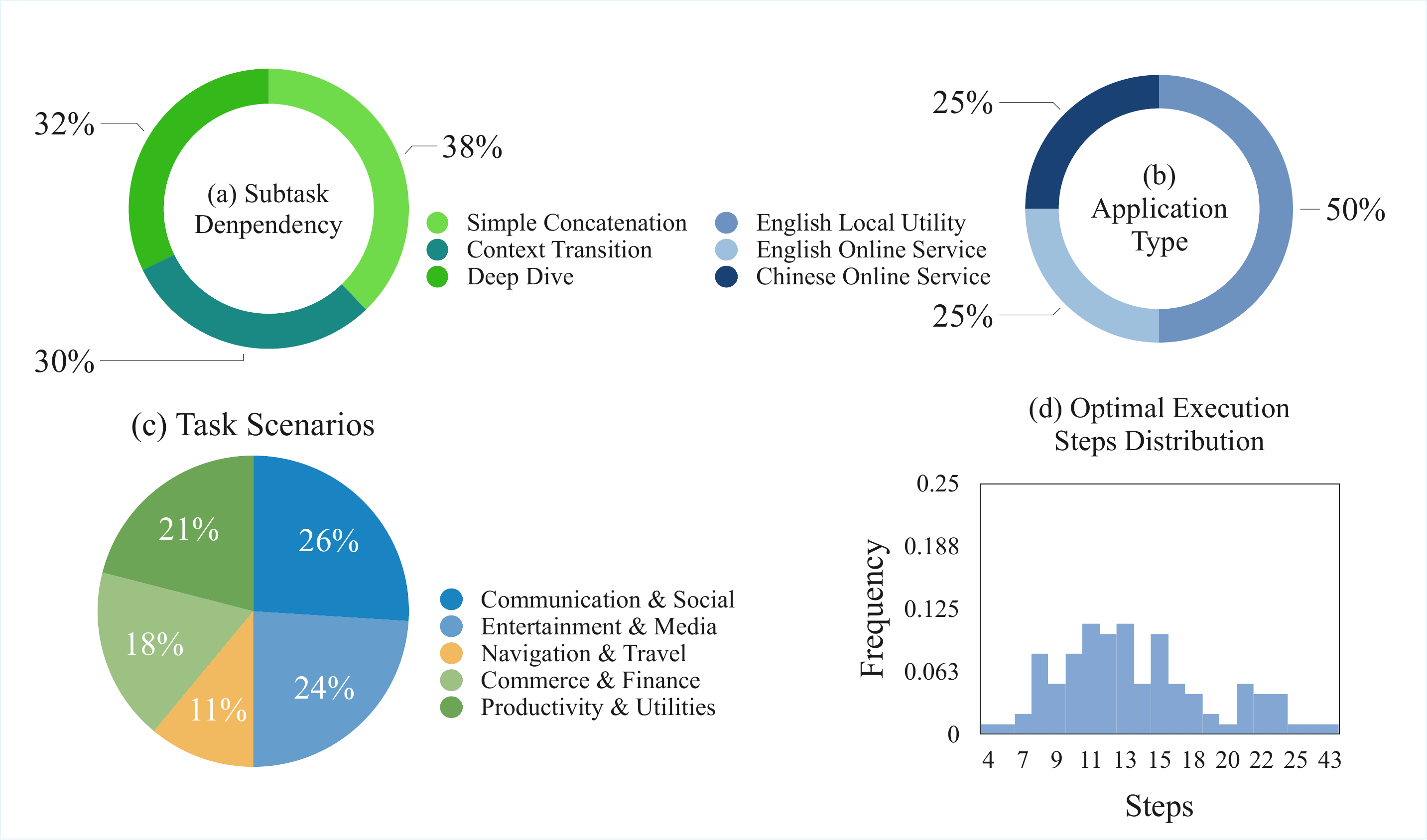}
    \caption{Data statistics of \benchmark instruction templates. There are 100 challenging templates in total, covering systematic subtask dependency structures, application types and using scenarios.}
    \label{fig:data-statistics}
    \vspace{-5mm}
\end{figure*}

Table \ref{tab:comparison} demonstrates the comparison with other existing GUI agent benchmarks. \benchmark features diverse application coverage, systematic investigation into long-horizon, compositional tasks and scalable interactive testing environments. Figure \ref{fig:data-statistics} demonstrates detailed benchmark task statistics and visualization.

\definecolor{c5}{RGB}{51,114,202}
\definecolor{c6}{RGB}{121,43,166}
\begin{figure*}[htbp]
\scriptsize
    \centering
    \includegraphics[width=0.90\textwidth]{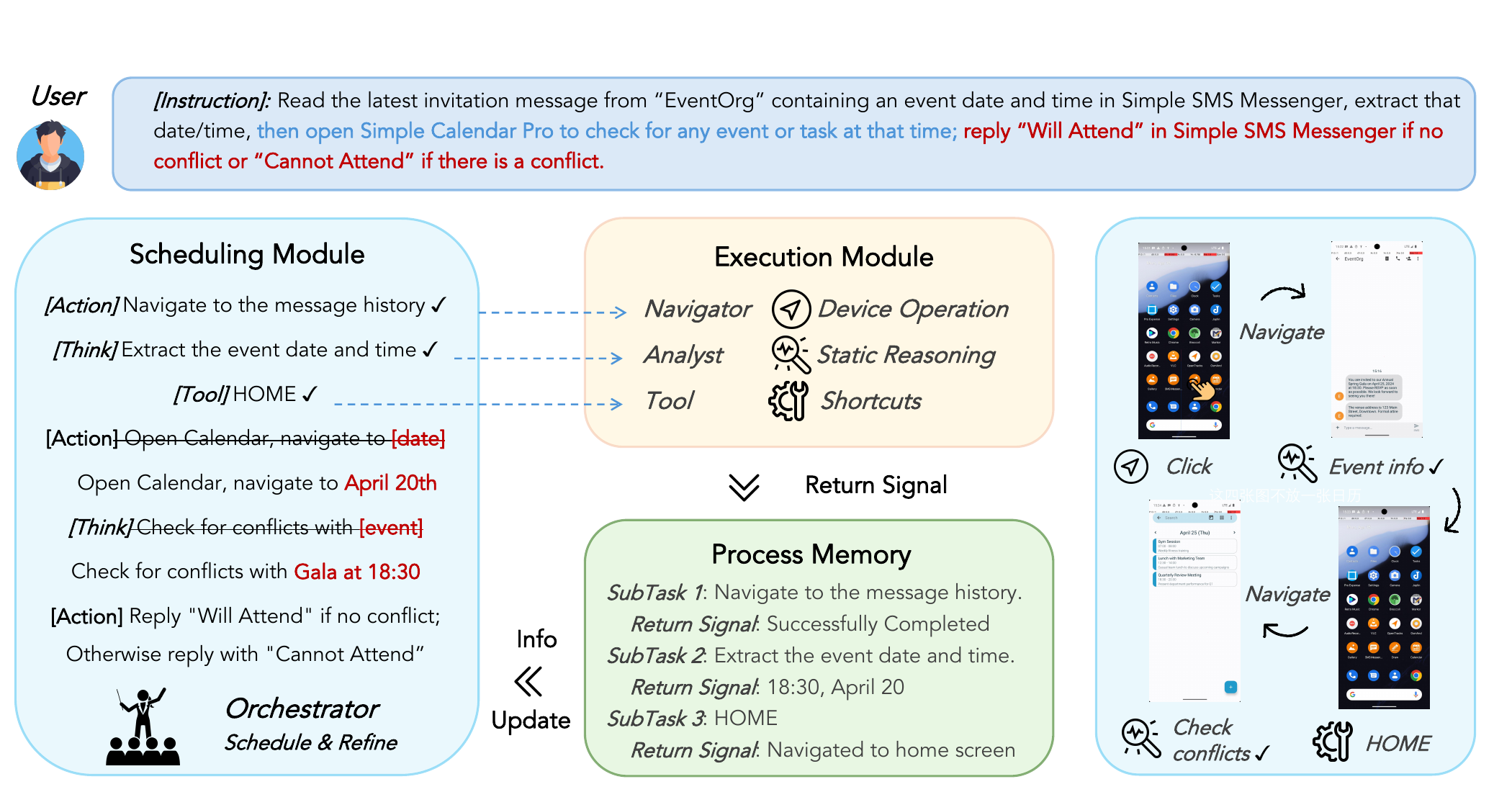}
    \caption{Overview of \agent framework. \agent follows a hierarchical adaptive orchestration design, with three interconnected modules Scheduling Layer, Execution Layer an Process Memory. We demonstrate a representative example where the different components seamlessly collaborate to retrieve invitation information, check for schedule conflicts, and reply accordingly.}
    \label{fig:agent-nexus-illustration}
    \vspace{-5mm}
\end{figure*}

\section{\agent: Systematic Scheduling for Compositional Task Automation}

To bridge the atomic-to-compositional generalization gap, we propose \agent, a lightweight and efficient scheduling system for tackling compositional mobile tasks.

As illustrated in Figure \ref{fig:agent-nexus-illustration}, \agent implements a \textbf{Scheduling Module} and \textbf{Execution Module} for collaborative task automation, and applies \textbf{Process Memory} to store and utilize intermediate information across subtasks. The scheduling module dynamically decomposes long-horizon compositional tasks into a series of self-contained atomic subtasks and assigns them to the execution module. Within the execution module, three key components collaborate: the \textbf{Analyst} handles general interface analysis and reasoning functionality using general large language models like GPT-4o; the \textbf{Navigator} manages device interactions using established mobile agents; and \textbf{Tool} executes predefined operations such as returning to the home screen. The process memory and subtask decomposition are dynamically updated after the return signal of each subtask, allowing effective adaptation to execution outcomes and managing dependencies between subtasks.
The detailed formulation of the working mechanism of \agent is elaborated in Appendix \ref{appendix:agent-nexus-formulation}.

\agent features a context reduction~\citep{teng2025atom} philosophy, where only a self-contained simple instruction is assigned to the execution module at each global step to significantly lower the cognitive load and alleviate the atomic-to-compositional generalization gap. With a systematic design, \agent goes beyond simple subtask planning to asynchronous information processing, management, and task instantiation. Different from previous mobile agent workflows \citep{wang2024mobile2,wang2025mobilee} that invoke the backbone model (GPT-4o) for potential information memorization at each step, \agent employs macro-level scheduling with multi-agent collaboration in the execution module, resulting in controlled computational redundancy. With plug-and-play feature of \agent, any developed low-level task execution agent can be conveniently integrated into our system.

\section{Experiments}
\label{sec::exp}

\subsection{Setup}
\label{sec:experimental-setup}
We evaluate representative mobile agents implemented as agentic workflow and agent-as-a-model.

Agentic workflow baselines include (i) M3A~\citep{rawles2024androidworld}: multimodal agent for Android integrating ReAct-style~\citep{yao2023react} and Reflexion-style~\citep{shinn2023reflexion} prompting; (ii) Mobile-Agent-v2~\citep{wang2024mobile2}: A multi-agent architecture for mobile device operation assistance with planning agent, decision agent, and reflection agent and memory module; (iii) Mobile-Agent-E~\citep{wang2025mobilee}: a hierarchical multi-agent framework capable of solving complex tasks by sub-goal breakdown and self-evolution through past experience. 

Agent-as-a-model baselines include (i) OS-Atlas-7B-Pro~\citep{wu2024atlas}: OS-Atlas-7B-Pro is a foundation action model for GUI with large-scale pre-training on 13 million grounding training data and fine-tuning on massive trajectory data. It is developed from QWen2-VL-7B; (ii) UI-TARS-7B-SFT~\citep{qin2025ui}: UI-TARS are a series of native GUI agents powered by domain-specific pretraining, action fine-tuning, and agent DPO. Following the authors' recommendation and preliminary experiments, we choose the SFT versions for mobile test scenarios. UI-TARS-7B-SFT is developed on QWen2-VL-7B.

Apart from the above-mentioned 5 baselines, we also test \agent on the \benchmark benchmark. We use GPT-4o as the backbone model of orchestrator in the scheduling module and analyst in the execution module, and try plugging both M3A model and UI-TARS-7B-SFT as the navigator in the execution module.
More implementation details are elaborated in Appendix \ref{appendix:implementation-details}.

\subsection{Main Results}
Detailed experimental results of tested agents on all three subsets of \benchmark are listed in Table \ref{tab:results-offline}, Table and Table \ref{tab:results-online-merged}. The detailed statistics of termination reasons on online service app tasks can be found in Appendix 
\ref{appendix:online-results}. The statistic information of the reasoning cost of the agents is listed in Table \ref{tab:inference-efficiency}. Here are overall findings:

(i) \benchmark poses substantial challenges on all five mobile agent baselines, with no agent exceeding task success rate of more than 50\% on all subsets.

(ii) Agentic workflow baselines show significantly better robustness than the SoTA agent-as-a-model baselines. But the inference time and token cost of GPT-4o-based workflows are remarkably higher.

(iii) Our \agent has effectively ignited the ability of mobile agents to handle the compositional tasks with controllable sacrifice of cost.

\begin{table}[!t]
\centering
\setlength{\tabcolsep}{3pt}
\resizebox{\textwidth}{!}{
\begin{tabular}{lcccccc}
\toprule
\multirow{2}{*}{Agent} & \multirow{2}{*}{\begin{tabular}[c]{@{}c@{}}Success\\ Rate\end{tabular}} & \multicolumn{5}{c}{Termination Reason} \\
\cmidrule(lr){3-7}
 &  & Successful & Premature & Budget Exceeded & Deemed Impossible & Collapse \\
\midrule
\multicolumn{7}{c}{\textit{Agentic Workflow (GPT-4o)}} \\
\midrule
M3A & 50.0 & 50.0 & 34.0 & 16.0 & 0.0 & 0.0 \\
Mobile-Agent-v2 & 30.0 & 30.0 & 34.0 & 34.0 & 0.0 & 2.0 \\
Mobile-Agent-E & 26.0 & 26.0 & 36.0 & 8.0 & 30.0 & 0.0 \\
\midrule
\multicolumn{7}{c}{\textit{Agent-as-a-Model}} \\
\midrule
OS-Atlas-7B-Pro & 2.0 & 2.0 & 20.0 & 72.0 & 0.0 & 6.0 \\
UI-TARS-7B-SFT & 6.0 & 6.0 & 8.0 & 84.0 & 2.0 & 0.0 \\
\midrule
\multicolumn{7}{c}{\textit{Ours}} \\
\midrule
\agent w/ M3A & \textbf{74.0} & 74.0 & 16.0 & 10.0 & 0.0 & 0.0 \\
\agent w/ UI-TARS-7B-SFT & 46.0 & 46.0 & 10.0 & 44.0 & 0.0 & 0.0 \\
\bottomrule
\end{tabular}
}
\vspace{0.3cm}
\caption{Task performance on the 50 tasks on local utility mobile apps (\offlinebench subset).}
\label{tab:results-offline}
\end{table}

\begin{table*}[!t]
\centering
\begin{minipage}[t]{0.48\textwidth}
\centering
\setlength{\tabcolsep}{2pt}
\caption{Inference efficiency (latency and cost per step) across agent variants.}
\label{tab:inference-efficiency}
\resizebox{\linewidth}{!}
{
\begin{tabular}{lcc}
\toprule
Agent & \begin{tabular}[c]{@{}c@{}}Inference Latency\\ (sec/step)\end{tabular} & \begin{tabular}[c]{@{}c@{}}Inference Cost\\ (USD/step)\end{tabular} \\
\midrule
\multicolumn{3}{c}{\textit{Agentic Workflow (GPT-4o)}} \\
\midrule
M3A & 14.77 & 0.037 \\
Mobile-Agent-v2 & 34.76 & 0.038 \\
Mobile-Agent-E & 38.20 & 0.037 \\
\midrule
\multicolumn{3}{c}{\textit{Agent-as-a-Model}} \\
\midrule
OS-Atlas-7B-Pro & 0.84 & 0.00047 \\
UI-TARS-7B-SFT & 4.35 & 0.0025 \\
\midrule
\multicolumn{3}{c}{\textit{Ours}} \\
\midrule
\agent w/ M3A & 18.86 & 0.040 \\
\agent w/ UI-TARS-7B-SFT & 6.53 & 0.0063 \\
\bottomrule
\end{tabular}
}
\end{minipage}
\hfill
\begin{minipage}[t]{0.48\textwidth}
\centering
\centering
\setlength{\tabcolsep}{2pt}
\caption{Success rates on English and Chinese online service app tasks.}
\label{tab:results-online-merged}
\resizebox{\linewidth}{!}{
\begin{tabular}{lcc}
\toprule
\multirow{2}{*}{Agent} & English Apps & Chinese Apps \\
 & \begin{tabular}[c]{@{}c@{}}Success Rate\end{tabular} & \begin{tabular}[c]{@{}c@{}}Success Rate\end{tabular} \\
\midrule
\multicolumn{3}{c}{\textit{Agentic Workflow (GPT-4o)}} \\
\midrule
M3A & 32.0 & 4.0 \\
Mobile-Agent-v2 & 12.0 & 12.0 \\
Mobile-Agent-E & 28.0 & 24.0 \\
\midrule
\multicolumn{3}{c}{\textit{Agent-as-a-Model}} \\
\midrule
OS-Atlas-7B-Pro & 4.0 & 4.0 \\
UI-TARS-7B-SFT & 8.0 & 8.0 \\
\midrule
\multicolumn{3}{c}{\textit{Ours}} \\
\midrule
\agent w/ UI-TARS-7B-SFT & 28.0 & 32.0 \\
\bottomrule
\end{tabular}
}
\end{minipage}
\end{table*}

\subsection{Analysis}

\subsubsection{Atomic-to-Compositional Generalization Gap}
We further investigate into the \textbf{Atomic-to-Compositional Generalization Gap} of mobile agents.

We select compositional instructions from the Simple Concatenation and Context Transition types in the \offlinebench subset, and manually decompose them into optimal atomic subtasks to study the gap of generalizing from atomic to compositional tasks. Inspired by Weak-to-Strong Generalization~\citep{burns2023weak}, we define the \textbf{Performance Gap Recovered (PGR)} as:

\begin{equation}
\text{PGR} = \frac{\text{Atomic-to-compositional performance} - \text{Weak performance}}{\text{Strong ceiling performance} - \text{Weak performance}}
\end{equation}

Here, \emph{weak performance} refers to executing the full compositional instruction directly, \emph{strong ceiling performance} assumes oracle assistance in task decomposition and context maintenance, and \emph{atomic-to-compositional performance}  measures how to which extent the gap is recovered. The results are shown in Table \ref{tab:generalization-gap}.

\begin{table}[!t]
\centering
\resizebox{\textwidth}{!}{
\begin{tabular}{lccccccc}
\toprule
\textbf{Agent} & SC-Comp & SC-Atom & CT-Comp & CT-Atom & Overall-Comp & Overall-Atom & Overall-PGR \\
\midrule
M3A & 55.0 & 70.0 & 60.0 & 93.0 & 57.0 & 80.0 ($\uparrow$87\%) & -- \\
Mobile-Agent-v2 & 40.0 & 45.0 & 27.0 & 47.0 & 34.0 & 46.0 ($\uparrow$33\%) & -- \\
Mobile-Agent-E & 35.0 & 45.0 & 27.0 & 47.0 & 31.0 & 46.0 ($\uparrow$45\%) & -- \\
OS-Atlas-7B-Pro & 5.0 & 0.0 & 0.0 & 13.0 & 3.0 & 6.0 ($\uparrow$100\%) & -- \\
UI-TARS-7B-SFT & 10.0 & 45.0 & 13.0 & 80.0 & 11.0 & 60.0 ($\uparrow$452\%) & -- \\
Agent-NEXUS w/ M3A & 70.0 & -- & 87.0 & -- & 77.0 & -- & 88.0 \\
Agent-NEXUS w/ UI-TARS-7B-SFT & 50.0 & -- & 73.0 & -- & 49.0 & -- & 76.0 \\
\bottomrule
\end{tabular}
}
\vspace{0.3cm}
\caption{Atomic-to-Compositional Generalization Gap for tested mobile agents. SC refers to Simple Concatenation tasks, CT refers to Context Transition tasks. "-Comp'' is the performance when directly provided with compositional task instructions (Weak Performance), while "-Atom'' refers to Strong Ceiling with optimized subtask decomposition.}
\label{tab:generalization-gap}
\end{table}

From Table \ref{tab:generalization-gap}, we can see all tested agents perform notably better when provided with optimized atomic tasks, which reveals the challenges of how to manage sophisticated subtask composition in task instructions. The gap for UI-TARS-7B-SFT is especially significant. It is because UI-TARS-7B-SFT is intensively trained for GUI operation logic, but the limited memory span due to the agent-as-a-model nature and less-developed long horizon ability significantly restrains their portential for compositional tasks. OS-Atlas exhibit impressive performance on many action grounding and offline evaluation benchmarks~\citep{wu2024atlas}, but it frequently fails to recognize the current state and stop or switch after the successful completion of a subtask, resulting in low performance in both atomic and compositional end-to-end evaluation.
Multiple case study examples on why mobile agents fail with compositional tasks are available in Appendix \ref{appendix:error-analysis}.
\agent achieves significant performance improvements with PGR of 88\% and 76\%, which shows the effectiveness of our scheduling system.

\subsection{Futher Discussions}

In this section, we further discuss the performance variance across application types, agent architecture and information management mechanism.

Comparing Table \ref{tab:results-offline} and Table \ref{tab:results-online-merged}, we find that online service apps are more challenging for mobile agents than local utility apps, even with similar task templates. This can be attributed to (i) more fancy UI designs, (ii) less interface maintain (for example, multiple Chinese online service apps do not support accessibility tree acquisition well, which significantly hinder the performance of M3A agent), and (iii) more distractions~\citep{ma2024caution}, e.g., advertisement in PicsArt and search recommendations.

In terms of agent architecture, agentic workflows outperform agent-as-a-model systems on complex tasks due to their structured reasoning and collaborative prompt design. However, this comes with higher inference cost and latency (Table~\ref{tab:inference-efficiency}). Agent-as-a-model solutions are efficient, showing promise for real-world deployment, but struggle with compound logic like conditional branches.

Memory management is a key factor for system-level mobile agent. We observe three patterns: \textit{no memory} (e.g., OS-Atlas), \textit{partial memory} (e.g., UI-TARS with short screenshot history), and \textit{proactive memory} (e.g., M3A with self-reflection or Mobile-Agent-V2 with step-wise note-taking). While proactive memory improves performance on context-dependent tasks, it introduces redundant computation and cognitive load.

Our proposed \agent strikes a balance by introducing a decoupled high-level scheduler that dynamically allocates subtasks and orchestrates navigation, perception, and reasoning. This modular design improves both adaptability and efficiency, enabling robust handling of long-horizon, compound tasks.

\section{Conclusion}
In this work, we present a comprehensive investigation into the atomic-to-compositional generalization challenge for mobile agents. To this end, we introduce \benchmark, a benchmark that systematically evaluates agent performance across three categories of subtask dependency structure and diverse real-world mobile apps. Extensive experiments reveal that existing agents struggle with long-horizon progress management, intermediate information transition, and reasoning-action integration. To address these challenges, we propose \agent, a lightweight and modular scheduling system that orchestrates hierarchical planning with decoupled execution. Our system significantly improves compositional task performance while maintaining efficiency. Together, \benchmark and \agent provide a unified foundation for developing and evaluating next-generation system-level intelligent agents capable of robust mobile interaction in complex environments.

\bibliographystyle{unsrt}
\bibliography{neurips_2025}


\appendix

\section{Limitations}
\label{appendix:limitations}
We acknowledge two primary limitations in our study. First, due to the ever-changing nature of online service interfaces, experimental environments on applications such as RedNote and YouTube may slightly differ across time. To mitigate this, we adopt the following strategies: (i) we include a benchmark subset focusing on local utility apps, \offlinebench, to provide a stable and controllable evaluation platform, while also incorporating bilingual online service apps to ensure broad coverage of real-world scenarios; (ii) for online service apps that lack direct interfaces for internal state control, we manually configure the test environments to ensure fairness in evaluation.
Second, our experiments are conducted solely on Android devices. While this choice is justified by the openness and flexibility of the Android ecosystem for agent development and control, it may slightly limit the generalizability of our findings to other mobile platforms such as iOS.

\section{Ethic Statement}
\label{appendix:ethics}

Our work contributes to the development of more capable and reliable mobile GUI agents through the design of a benchmark \benchmark and a modular scheduling system \agent for compositional task automation. These advancements have the potential to improve user accessibility, reduce digital operation barriers, and empower individuals with limited technical literacy to interact more effectively with mobile services.

In particular, intelligent mobile agents can assist users with visual impairments, automate repetitive app-based workflows, and enhance productivity across communication, finance, and navigation scenarios. However, such systems may also raise concerns around excessive automation, user autonomy, and behavioral manipulation if deployed without transparent boundaries or consent mechanisms.

We explicitly caution against deploying the provided system or datasets in applications involving sensitive personal data, opaque surveillance, or manipulative behavior design. We encourage future developers to prioritize user agency, build in clear feedback loops, and comply with ethical standards when extending this work into production-grade systems.

\section{Implementation Details}
\label{appendix:implementation-details}
For the mobile agents implemented as agentic workflow, we use GPT-4o~\citep{openai2024gpt4o} as the backbone model. We set temperature of backbone models to 0.0 for better reproducibility. The maximum
length of the input sequence is 4096 tokens.

Notably, the mobile agents implemented as agentic workflow, such as Mobile-Agent-v2, require multiple API calls per step, often exceeding the rate limit of GPT-4o during end-to-end task execution. To ensure that the experimental results are not influenced by user plan rate limits, we introduced a pause between each step of task execution. Consequently, the latency in real-world implementations may be longer than the results reported in this paper.

For OS-Atlas-4B-Pro and OS-Atlas-7B-Pro, we prompt the models to generate a brief thought describing the next action followed by the action specification every step, following the official HuggingFace model card~\citep{OS-Atlas-Pro-7B}.

UI-TARS-7B-SFT implement a short-term memory mechanism by providing previous N screenshots in every steo input (N is set as a hyper-parameter). We set N to 5 following the suggested default setting in official repository~\citep{UI-TARS}.

For experiments in local utility apps (evaluation on \offlinebench subset), we conduct experiments in Android Emulators and adopt the same system version of AndroidWorld~\citep{rawles2024androidworld}. The hardware choice is Pixel 6, the System Image is Tiramisu, API Level 33. And for evaluations on online service apps, we set up experimental environment on physical devices to avoid restrictions and censorship of account login on emulators.

We use adbkeyboard instead of the default keyboard on Android devices to support Chinese characters input and the specific requirements of Mobile-Agent-V2 implementation. 

We standardize the method of launching target apps by tapping their icons on the home screen. Accordingly, we remove the open\_app action from the action space of OS-Atlas-Pro-7B and M3A, and ensure that all required apps are placed on the device's home screen during the testing of each subset.

In the execution module of \agent, we currently apply a simplified implementation, where the only tool in HOME.

\section{Error Analysis}
\label{appendix:error-analysis}
\begin{figure*}[t]
\scriptsize
    \centering
    \includegraphics[width=0.95\textwidth]{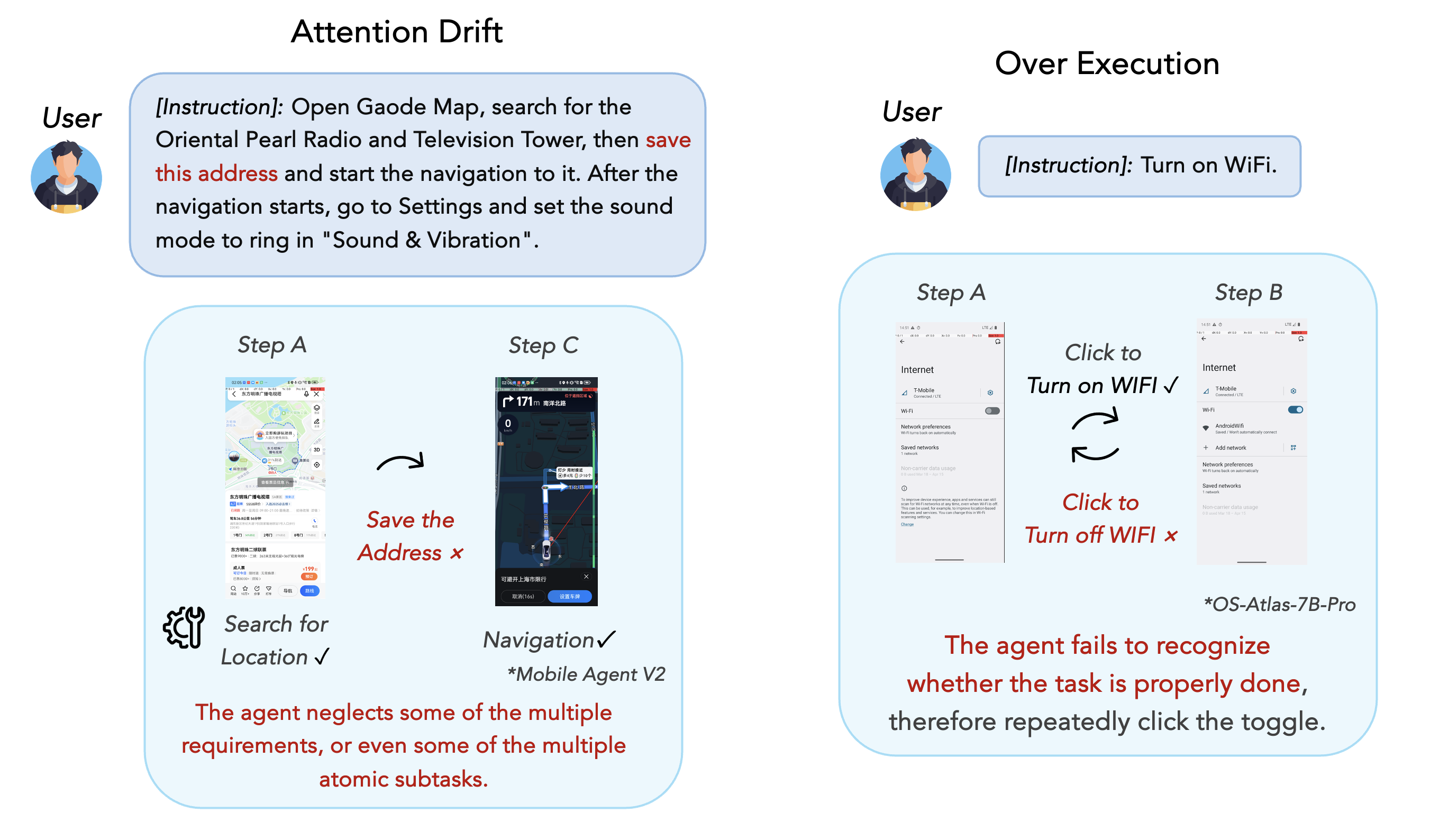}
    \caption{Representative examples of attention-drift and over-execution failure mode.}
    \label{fig:failure_mode1}
    \vspace{-5mm}
\end{figure*}

\begin{figure*}[t]
\scriptsize
    \centering
    \includegraphics[width=0.95\textwidth]{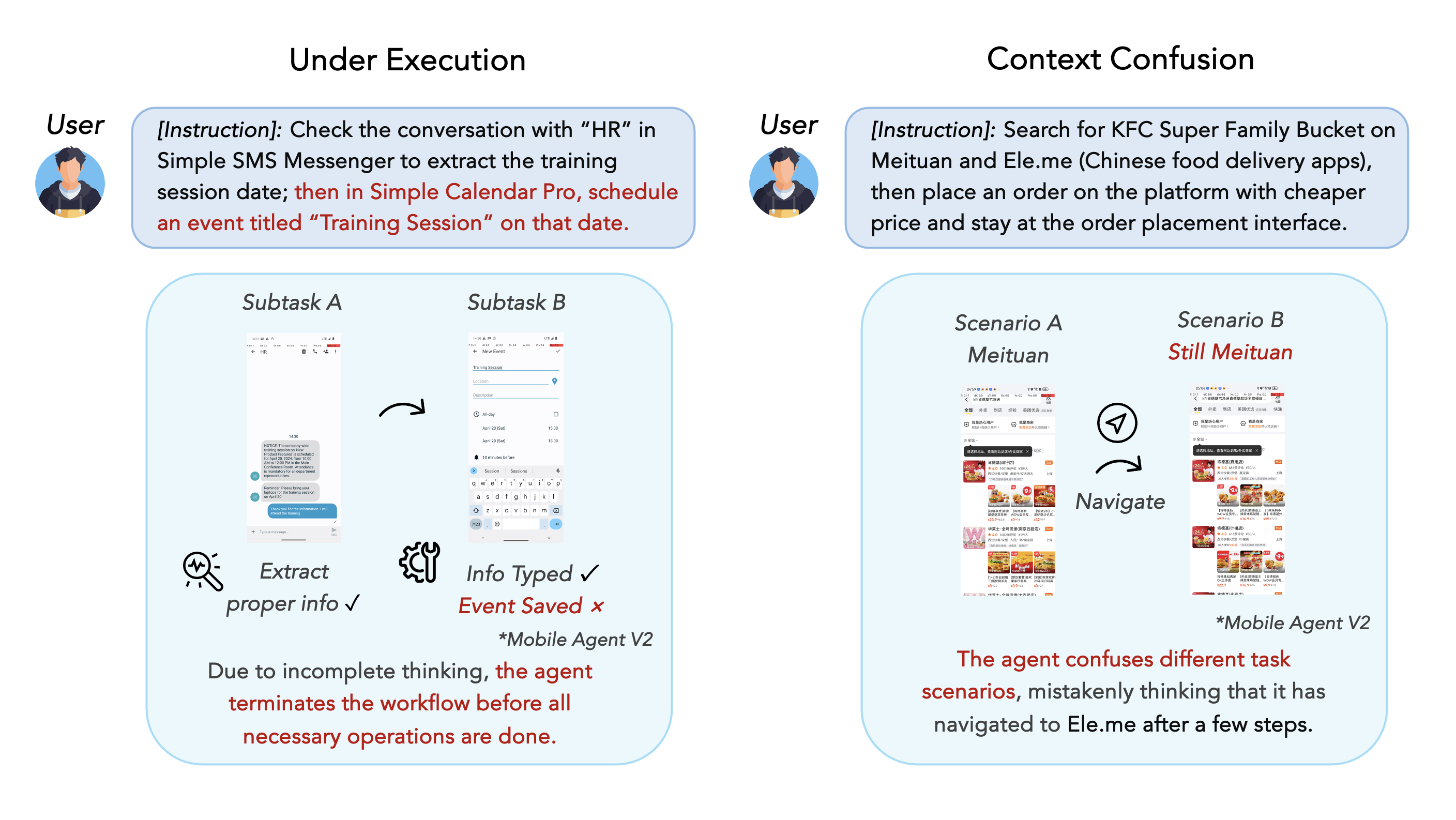}
    \caption{Representative examples of under-execution and context confusion failure mode.}
    \label{fig:failure_mode2}
    \vspace{-5mm}
\end{figure*}

\begin{figure*}[t]
\scriptsize
    \centering
    \includegraphics[width=0.95\textwidth]{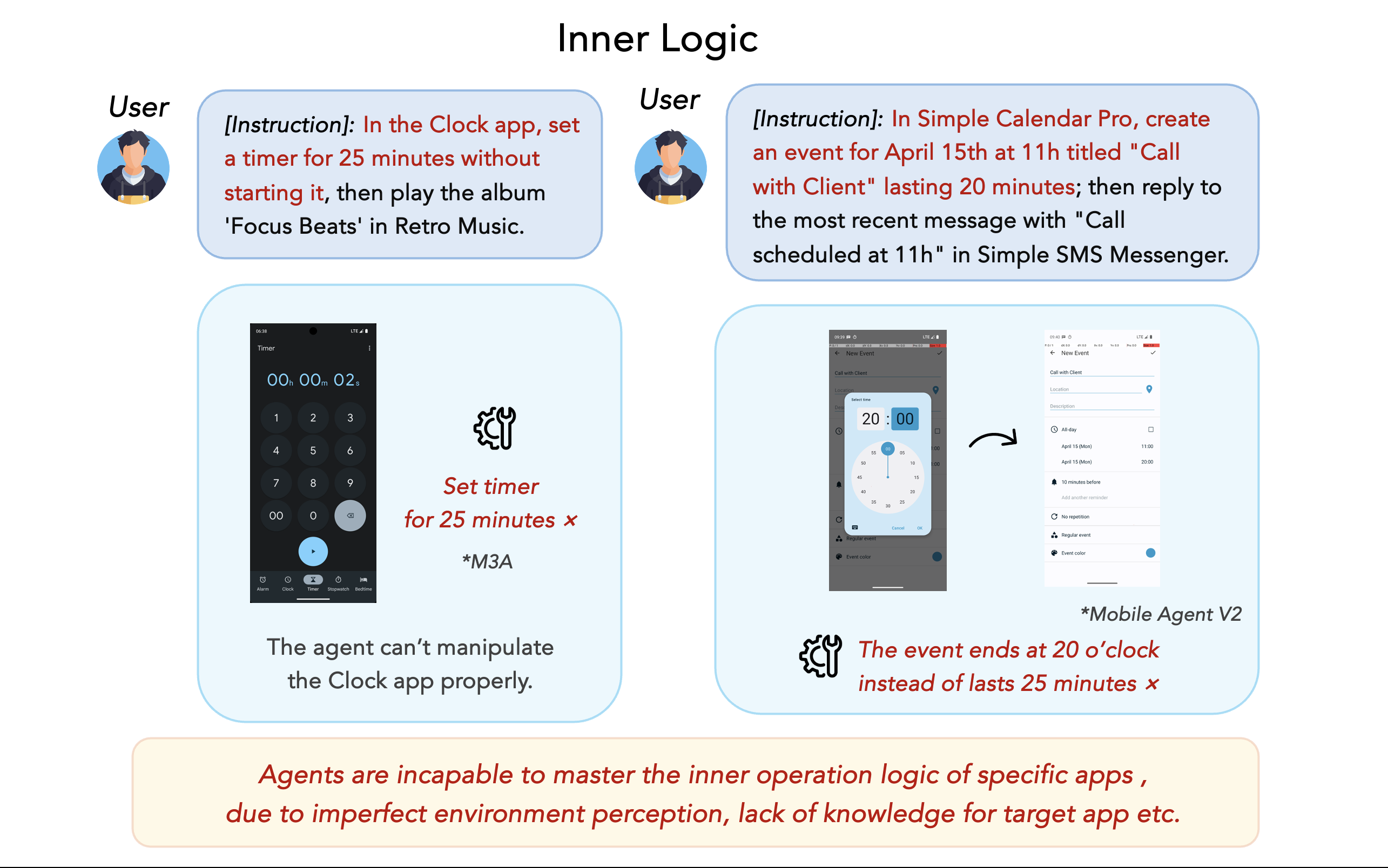}
    \caption{Representative examples of inner logic error failure mode.}
    \label{fig:failure_mode3}
    \vspace{-5mm}
\end{figure*}

In evaluation experiments, we have observed that existing mobile agents exhibit some types of representative failure modes. We provide some typical error cases across different tested agents in Figure \ref{fig:failure_mode1}, Figure \ref{fig:failure_mode2} and Figure \ref{fig:failure_mode3}. Some representative failure modes of mobile agents on compositional tasks include:

\begin{itemize}
    \item \textbf{Over-execution:} Over-execution arises when the agent issues \emph{redundant or irrelevant UI actions that go beyond the minimal sequence required to satisfy the instruction}.  These actions are syntactically valid clicks, scrolls, or text entries, yet they are \emph{semantically unjustified}: the task is already complete, the necessary intermediate data are missing, or the interface is meant for deliberation rather than interaction.  We identify three recurrent variants:

    \begin{itemize}
        \item \textbf{Deficient progress monitoring.} Without reliable completion detection, the agent fails to recognize task termination and issues gratuitous post-completion actions. One representative mistake of this type is when OS-Atlas is conducting device-setting tasks like \textit{`Turn on the bluetooth'}, it may repeatedly click the toggle and turn it on and off.
        \item \textbf{Faulty information management in context transitions.} When intermediate data fail to propagate, the agent nonetheless advances, injecting invented values that diverge from the intended workflow.
        \item \textbf{Breakdown of thinking-acting arbitration.} In Context Transition or Deep Dive tasks that require alternation between general reasoning (like arithmetic calculation and text summarization) and low‑level execution, the agent remains in the action loop instead of pausing for reasoning, generating purposeless interactions that do not meaningfully change the state.
    \end{itemize}

    \item \textbf{Under-execution:} Under-execution denotes the premature termination of a subtask or the entire workflow before all mandatory operations are fulfilled. Conceptually, it parallels the \emph{under‑thinking} phenomenon reported for reasoning models~\citep{wang2025thoughts}, in which a promising line of reasoning is truncated. This failure mode is frequently observed in UI-TARS, where a limited working memory span and inadequate milestone recognition give rise to an oscillatory “ping-pong’’ pattern: the agent executes a few actions in Subtask A, abruptly switches to Subtask B, and keeps alternating without completing either.

    \item \textbf{attention drift: }Agent may neglect some of the multiple requirements, or even some of the multiple atomic subtasks within the compositional task instruction.
    \item \textbf{context confusion: }When directly provided with compositional task instructions, the agent may confuse different task scenarios and inter-scenario requirements.
    \item \textbf{switching failure: }Most of the tasks \benchmark involve more than one app. Some tested mobile GUI agents are just optimized for single app operations and struggle to switch between different apps. For example, the Open\_App action of Mobile-Agent-V2 only allows opening apps whose names are visible on the current screen. But the agent may repeatedly try this action when it has just finished the task within one app and hasn't switched to the home screen first.
    \item \textbf{Greedy information collection:} For tasks that involve gathering and processing information, certain agents display a \emph{greedy} strategy: they consult only the first page of a document, a thumbnail of an information-dense image, or the compact ledger preview on the Expense Pro home screen, ignoring the richer content available beyond these minimal views.
    \item \textbf{CoT inconsistency: }The tested baselines typically include thoughts and actions in each round's output. The thought part often includes a reasoning process that induces the next action to take. However, we find that the agent (especially for fine-tuned specialist agents like OS-Atlas-Pro and UI-TARS) 
    may generate mismatched thought and action in its output. This kind of reasoning-and-answer inconsistency matches some previous works on CoT faithfulness~\citep{turpin2023language,lanham2023measuring} and calls for more sophisticated training and prompting recipes to make better use of Chain-of-Thought reasoning in agent tasks.
    \item \textbf{inner operation logic: }A large proportion of the failure still results from the agents' incapability to handle the inner operation logic of the specific apps. This can be attributed to the imperfect environment perception, lack of knowledge for target app and suboptimal run-time reasoning and self-correction.
\end{itemize}

\section{Evaluation Metric Details}
\label{appendix:evaluation_metrics}
Here we provide more detailed information on the calculation of some of the evaluation metrics.

\paragraph{Termination Reason}
In the evaluation experiments of \benchmark, we set the maximum step budget for each task to approximately twice the number of optimized steps annotated by humans. If the agent fails to emit a stop signal within this budget, the task is forcibly terminated and marked as \textit{Step Budget Exceeded}.

Mobile-Agent-E implements an early-stop mechanism, where repetitive same operations and repetitive error operations (the reflection agent deems the operation doesn't meet expectation) will result in task termination. We categorize this kind of task termination as Deemed Impossible.

\paragraph{Inference Latency}
We calculate the average inference time per step to measure the temporal efficiency of the tested mobile agents. To avoid the affects of external noises, we only calculated the time spent on reasoning (api response latency for proprietary models and inference latency for locally deployed models) rather than the overall time interval between the steps.

For mobile agents that are implemented as agent-as-a-model (OS-Atlas-7B-Pro and UI-TARS-7B-SFT), we deploy them on 1 NVIDIA A100-SXM4-80GB GPU with VLLM~\citep{kwon2023efficient} framework, and record the inference latency for each step.

\paragraph{Inference Cost}
All inference costs are reported using the publicly available prices from April~2025.
GPT-4o is priced at \$2.50 per million input tokens and \$10.00 per million output tokens.
Qwen-VL-Plus, employed as the icon-captioning model in
Mobile-Agent-V2~\citep{wang2024mobile2} and Mobile-Agent-E~\citep{wang2025mobilee},
is priced at \$0.21 per million input tokens and \$0.63 per million output tokens.
For specialist agents, we follow the approximation strategy of Aguvis~\citep{xu2024aguvis}.
Specifically, we use the API prices of their underlying foundation models offered by
third-party providers: for OS-Atlas-7B-Pro and UI-TARS-7B-SFT
(both based on Qwen2-VL-7B), we apply the
OpenRouter\footnote{\url{https://openrouter.ai/models}} rate of
\$0.20 per million input tokens and \$0.20 per million output tokens.

\section{\agent Formulation}
\label{appendix:agent-nexus-formulation}
\subsection{Hierarchical Planning Formulation}
Building upon the compositional task definition $\mathcal{T}{\text{comp}} \triangleq \langle \mathcal{A}{\text{sub}}, \mathcal{D} \rangle$, we formalize our hierarchical planning approach as $\langle \mathcal{M}, \mathcal{E}, \mathcal{P} \rangle$, where:
\begin{itemize}
\item $\mathcal{M}$ represents the Scheduling Module
\item $\mathcal{E}$ represents the Execution Module
\item $\mathcal{P}$ represents the Process Memory
\end{itemize}
The Scheduling Module maps the current state, goal, and process memory to a plan:
\begin{equation}
\mathcal{M}: \mathcal{S} \times \mathcal{G} \times \mathcal{P} \rightarrow \Pi
\end{equation}
A plan consists of an ordered sequence of subtasks:
\begin{equation}
\pi = (t_1, t_2, \ldots, t_n)
\end{equation}
Each subtask $t_i$ is defined as:
\begin{equation}
t_i = \langle id_i, type_i, instr_i \rangle \text{ with } type_i \in \{act, think, tool\}
\end{equation}
\begin{itemize}
\item The $act$ type handles UI manipulation
\item The $think$ type manages reasoning without environment modification
\item The $tool$ type executes predefined operations like returning to home screen
\end{itemize}
\subsection{Execution Module}
This layer implements different execution strategies based on subtask type:
\textbf{For $act$ subtasks:}
\begin{equation}
\mathcal{E}{act}: \mathcal{S} \times instr \times H \rightarrow \mathcal{S}' \times \{steps\} \times \{0,1\}
\end{equation}
This function executes environment interactions within a maximum step limit $H$, returning the new state, execution history, and completion status.
\textbf{For $think$ subtasks:}
\begin{equation}
\mathcal{E}{think}: \mathcal{S} \times instr \rightarrow \mathcal{R}{think} \times \{0,1\} \times \{0,1\}
\end{equation}
This function performs reasoning on the current state, returning the reasoning result, completion status, and any failure flag.
\textbf{For $tool$ subtasks:}
\begin{equation}
\mathcal{E}{tool}: instr \rightarrow \mathcal{S}' \times \{0,1\}
\end{equation}
This function executes predefined operations, returning the new state and completion status.
\subsection{Process Memory}
The process memory $\mathcal{P}$ serves as a chronological record of executed subtasks and their outcomes, enabling context maintenance across planning cycles. For each executed subtask $t_i$, an entry is stored:
\begin{equation}
e_i = \langle i, instr_i, \mathcal{R}i \rangle
\end{equation}
Where:
\begin{itemize}
\item $i$ is the global subtask counter
\item $instr_i$ is the instruction
\item $\mathcal{R}i$ is the execution result that varies by type:
\begin{itemize}
\item $\mathcal{R}{act}$ contains completion status and detailed execution logs with reasoning, actions, and reflections for each step
\item $\mathcal{R}{think}$ includes reasoning text and any failure indicators
\item $\mathcal{R}{tool}$ contains operation results
\end{itemize}
\end{itemize}
This structure effectively captures intermediate information from both simple concatenation and context transition tasks, while the think subtasks specifically address the deep dive category requiring additional reasoning.

\subsection{Dynamic Re-planning}
After each subtask execution, the scheduling layer recomputes the plan based on updated process memory:
\begin{equation}
\pi_{t+1} = \mathcal{M}(s_{t+1}, g, \mathcal{P}{t+1})
\end{equation}
This mechanism allows \agent to:
\begin{itemize}
\item Adapt to execution outcomes
\item Leverage intermediate information
\item Handle dependencies between subtasks
\end{itemize}
The dynamic re-planning approach is particularly effective for context transition and deep dive compositional tasks, where subsequent subtasks depend on previous execution states or derived information. By maintaining detailed execution records and continuously reassessing the plan, the system can successfully bridge the gap between atomic and compositional task execution, providing a robust framework for addressing complex mobile interaction scenarios.
\subsection{Modular System Architecture}
As an attempt towards system-level edge intelligence, \agent features a modular, pluggable architecture that can effectively integrate various foundation models for device manipulation. This design allows seamless integration of different atomic task executors such as M3A and UI-TARS within the Execution Layer, where each implements the standardized interface $\mathcal{E}{act}$ while leveraging their specific strengths.
Similarly, the Scheduling Layer can utilize various LLMs, while maintaining consistent communication protocols through the Process Memory. This extensible approach offers several advantages:
\begin{itemize}
\item Enables rapid adaptation to emerging foundation models without architectural changes
\item Maximizes resource efficiency through model specialization
\item Facilitates flexible deployment across diverse mobile environments
\item Supports progressive system refinement by allowing individual component optimization while maintaining overall system integrity
\end{itemize}

\section{Detailed Results on Online Subsets}
\label{appendix:online-results}

The detailed statistics of performance-related experimental results on Chinese and English online service apps are demonstrated in Table \ref{tab:results-chinese-online-detail} and Table \ref{tab:results-english-online-detail}.

\begin{table}[!t]
\centering
\setlength{\tabcolsep}{3pt}
\resizebox{\textwidth}{!}{
\begin{tabular}{lcccccc}
\toprule
\multirow{2}{*}{Agent} & \multirow{2}{*}{\begin{tabular}[c]{@{}c@{}}Success\\ Rate\end{tabular}} & \multicolumn{5}{c}{Termination Reason} \\
\cmidrule(lr){3-7}
 &  & Successful & Premature & Budget Exceeded & Deemed Impossible & Collapse \\
\midrule
\multicolumn{7}{c}{\textit{Agentic Workflow (GPT-4o)}} \\
\midrule
M3A & 4.0 & 4.0 & 12.0 & 44.0 & 0.0 & 40.0 \\
Mobile-Agent-v2 & 12.0 & 12.0 & 28.0 & 60.0 & 0.0 & 0.0 \\
Mobile-Agent-E & 24.0 & 24.0 & 36.0 & 4.0 & 28.0 & 8.0 \\
\midrule
\multicolumn{7}{c}{\textit{Agent-as-a-Model}} \\
\midrule
OS-Atlas-7B-Pro & 4.0 & 4.0 & 8.0 & 88.0 & 0.0 & 0.0 \\
UI-TARS-7B-SFT & 8.0 & 8.0 & 24.0 & 68.0 & 0.0 & 0.0 \\
\midrule
\multicolumn{7}{c}{\textit{Ours}} \\
\midrule
\agent w/ UI-TARS-7B-SFT & 32.0 & 32.0 & 4.0 & 64.0 & 0.0 & 0.0 \\
\bottomrule
\end{tabular}
}
\vspace{0.3cm}
\caption{Task performance on interactive tests across Chinese online service mobile apps.}
\label{tab:results-chinese-online-detail}
\end{table}

\begin{table}[!t]
\centering
\setlength{\tabcolsep}{3pt}
\resizebox{\textwidth}{!}{
\begin{tabular}{lcccccc}
\toprule
\multirow{2}{*}{Agent} & \multirow{2}{*}{\begin{tabular}[c]{@{}c@{}}Success\\ Rate\end{tabular}} & \multicolumn{5}{c}{Termination Reason} \\
\cmidrule(lr){3-7}
 &  & Successful & Premature & Budget Exceeded & Deemed Impossible & Collapse \\
\midrule
\multicolumn{7}{c}{\textit{Agentic Workflow (GPT-4o)}} \\
\midrule
M3A & 32.0 & 32.0 & 0.0 & 68.0 & 0.0 & 0.0 \\
Mobile-Agent-v2 & 12.0 & 12.0 & 12.0 & 76.0 & 0.0 & 0.0 \\
Mobile-Agent-E & 28.0 & 28.0 & 24.0 & 16.0 & 28.0 & 4.0 \\
\midrule
\multicolumn{7}{c}{\textit{Agent-as-a-Model}} \\
\midrule
OS-Atlas-7B-Pro & 4.0 & 4.0 & 0.0 & 88.0 & 0.0 & 8.0 \\
UI-TARS-7B-SFT & 8.0 & 8.0 & 4.0 & 88.0 & 0.0 & 0.0 \\
\midrule
\multicolumn{7}{c}{\textit{Ours}} \\
\midrule
\agent w/ UI-TARS-7B-SFT & 28.0 & 28.0 & 12.0 & 60.0 & 0.0 & 0.0 \\
\bottomrule
\end{tabular}
}
\vspace{0.3cm}
\caption{Task performance on offline tests across English online service mobile apps.}
\label{tab:results-english-online-detail}
\end{table}

\section{Data Annotation Details}
\subsection{Optimized Operation Steps Annotation}
\begin{tcolorbox}[colback=gray!5!white, colframe=gray!75!black, title=Optimized Operation Steps Annotation Prompt, breakable]

You are asked to annotate the **optimized operation steps** for completing a given mobile task instruction.

Your available actions correspond to **basic atomic interactions** on a mobile device, such as:
\begin{itemize}
    \item Tapping a button or UI element
    \item Scrolling the screen
    \item Typing text into an input field
    \item Swiping or switching between tabs
    \item Confirming pop-ups or dialog boxes
\end{itemize}

Each of the above counts as **one step**, and you should only include steps that are strictly necessary to accomplish the task. At the end, you will return a numbered list of such steps, which represents the **shortest correct path** to complete the task.

To perform the annotation properly, follow these instructions:
\begin{itemize}
    \item Carefully read the given task instruction and identify the user's explicit goal.
    \item Interact with the mobile device (real or simulated) to complete the task using the **most efficient sequence** of operations.
    \item Avoid redundant actions, exploratory attempts, or trial-and-error. The steps should reflect a fluent and goal-driven interaction.
    \item Include all required intermediate actions, even if trivial (e.g., closing a pop-up, navigating back).
    \item Ensure the steps are listed in strict execution order and cover the entire process from the home screen to task completion.
\end{itemize}

\end{tcolorbox}

\subsection{Task Success Verification}

\begin{tcolorbox}[colback=gray!5!white, colframe=gray!75!black, title=Task Brainstorming Prompt, breakable]
To annotate whether a task execution is successful, follow the verification criteria below:

\begin{itemize}
    \item Begin by reviewing the original task instruction and the full action trajectory or execution video.
    \item Check if the final state of the interface clearly reflects that the user goal has been achieved (e.g., the correct app screen is reached, the expected content is visible, or the correct response is sent).
    \item Partial completions or approximate matches should \textbf{not} be labeled as successful. A task is only marked as \texttt{Successful} if \textbf{all explicit requirements} are satisfied.
    \item If any required subgoal is missed, the final state is ambiguous, or incorrect information is input or output, label the task as \texttt{Failed}.
    \item For ambiguous cases, provide justification in the annotation field and flag for reviewer inspection.
    \item Use binary labels: \texttt{Successful} or \texttt{Failed}.
\end{itemize}

\textbf{Example:} For the instruction “Send ‘I have joined the meeting’ to Yuan on WeChat after starting a Tencent Meeting session”:
\begin{itemize}
    \item \texttt{Successful} if the agent opens Tencent Meeting, starts a session, and then opens WeChat, enters the correct chat, and sends the correct message.
    \item \texttt{Failed} if the meeting is not started, the message is sent to the wrong contact, or no message is sent.
\end{itemize}
\end{tcolorbox}

\section{Data Collection}
\label{appendix:data-collection}
\subsection{Compound Logic Integration}
To ensure the diversity and coherence of our compositional task instructions, we incorporate templates that embody compound logic structures, including sequential, parallel, conditional, and hierarchical forms. We employ logical connectors such as ``if'', ``or'', ``then'', and ``define a subworkflow'' to express these structures clearly and consistently within the task templates. Table \ref{tab:compound-logic} presents formalized definitions of the four types of compound logic we adopt, following conventions from prior works~\citep{xu2024generalization, furuta2023exposing}.

\newpage
\begin{table}[!t]
\centering
\begin{tabular}{lll}
\toprule
\textbf{Logic Type} & \textbf{Notation} & \textbf{Description} \\
\midrule
Sequential  & \texttt{[a, b, c]}     & Subtasks must be completed in a specific order. \\
Conjunctive & \texttt{a \& b \& c}   & All components must be completed (logical AND). \\
Disjunctive & \texttt{a | b | c}     & Any one of the components is sufficient (logical OR). \\
Hierarchical & \texttt{Define(a $\rightarrow$ \{b, c\})} & Subtasks are grouped under a parent task with internal structure. \\
\bottomrule
\end{tabular}
\vspace{0.3cm}
\caption{Four types of compound logic used in task template construction and CheckPoint evaluation.}
\label{tab:compound-logic}
\end{table}

\subsection{Task Brainstorming}
After manually obtaining seed task instruction templates, we prompt large language models to brainstorm more task instructions, here is the prompt template:

\begin{tcolorbox}[colback=gray!5!white, colframe=gray!75!black, title=Task Brainstorming Prompt, breakable]

You are an expert in designing mobile task instructions for benchmarking intelligent agents. Based on the provided seed instructions, logic patterns, task types, and app list, please write \textbf{[num] diverse and realistic mobile task instructions} that reflect compositional logic. Ensure the tasks are interactive, grounded in real app capabilities, and distributed across different categories.

\textbf{Selected Apps:}  
\{apps\}

\textbf{Compositional Logic Types:}  

- Sequential: Do A, then B  

- Conjunctive: Do A and B  

- Disjunctive: Do A or B  

- Hierarchical: Define a task A that includes subtasks B and C  

\textbf{Compositional Task Types:}  

- \textbf{Simple Concatenation:} A sequence of subtasks \texttt{(a\textsubscript{1}, a\textsubscript{2}, ..., a\textsubscript{n})} that are executed independently without cross-subtask state dependencies.

- \textbf{Context Transition:} A composition where certain subtasks depend on the output states of others, i.e., the output of \texttt{a\textsubscript{i}} is used as input to \texttt{a\textsubscript{j}}.

- \textbf{Deep Dive:} A semantically dense task requiring exploration within a single app or context, often involving multiple dependent atomic actions.

\textbf{Seed Instructions (Examples):}  
\{seed\_instructions\}

\textbf{Guidelines:}  

- Each instruction should include at least one form of compound logic.  

- Use apps and scenarios from different categories such as \texttt{Communication}, \texttt{Navigation}, \texttt{Productivity}, etc.  

- Follow natural language tone and real-world feasibility.  

- Assume the agent has full control over the mobile interface and app environment.

\textbf{Output Format:}  
Task 1: [Instruction]  

Task 2: [Instruction]  

...  

Task 5: [Instruction]

\end{tcolorbox}

We instantiate this prompt with real-world app names and seed examples, then apply it to LLMs for task brainstorming. Generated instructions are then curated for balance and quality.

\end{document}